\documentclass{article}

    \PassOptionsToPackage{numbers, compress}{natbib}

    \usepackage[preprint]{neurips_2024}

\usepackage[utf8]{inputenc} 
\usepackage[T1]{fontenc}    
\usepackage{hyperref}       
\usepackage{url}            
\usepackage{amsfonts}       
\usepackage{nicefrac}       
\usepackage{microtype}      

\usepackage{amsmath}

\usepackage{float}
\usepackage{graphicx}      
\usepackage{booktabs}      
\usepackage{array}         
\usepackage{colortbl}      
\usepackage[table]{xcolor} 
\definecolor{navyblue}{rgb}{0.0, 0.0, 0.5}
\definecolor{oai-green-200}{RGB}{204, 255, 204}
\definecolor{oai-green-400}{RGB}{153, 255, 153}
\definecolor{oai-green-600}{RGB}{102, 255, 102}
\definecolor{oai-gray-300}{RGB}{200, 200, 200}
\definecolor{oai-gray-600}{RGB}{150, 150, 150}
\definecolor{orange!10}{RGB}{255, 230, 204}
\definecolor{yellow!10}{RGB}{255, 255, 204}
\usepackage{caption}       
\usepackage{multirow}      
\usepackage{rotating}      
\usepackage{pifont}    
\usepackage{arydshln}
\usepackage{xspace}
\usepackage{fontawesome}
\definecolor{yunze}{RGB}{255, 0, 0}  

\newcommand{\github}{\raisebox{-1.5pt}{\includegraphics[height=1.05em]{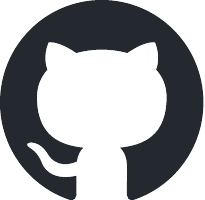}}\xspace}
\newcommand{\huggingface}{\raisebox{-1.5pt}{\includegraphics[height=1.05em]{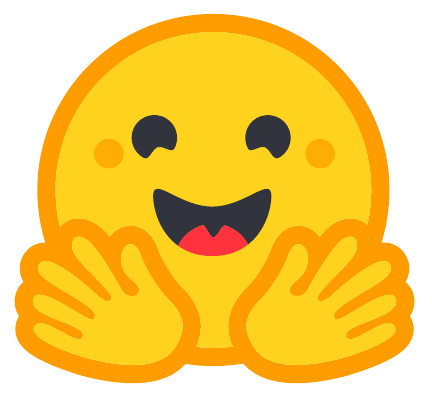}}\xspace}

\title{ST-Think: How Multimodal Large Language Models Reason About 4D Worlds from Ego-Centric Videos}

\author{%
  Peiran Wu$^{1}$\thanks{Equal contribution.}\quad
  Yunze Liu$^{2}$\footnotemark[1] \ \thanks{Project Leader.}\quad
  Miao Liu$^{3}$\quad
  Junxiao Shen$^{1,2}$\thanks{Corresponding Author.}\\[1ex]
  $^{1}$University of Bristol \quad
  $^{2}$X-Intelligence Labs \quad
  $^{3}$Meta\\[1ex]
  {\github\ \href{https://github.com/WPR001/Ego-ST}{\text{Evaluation Code}} \quad
   \huggingface\ \href{https://huggingface.co/datasets/openinterx/Ego-ST-bench}{\text{Ego-ST Bench}}
  }
}

\begin{document}

\maketitle


\begin{figure*}[htp]
    \vspace{-3em}
    \centering
    \includegraphics[width=\linewidth]{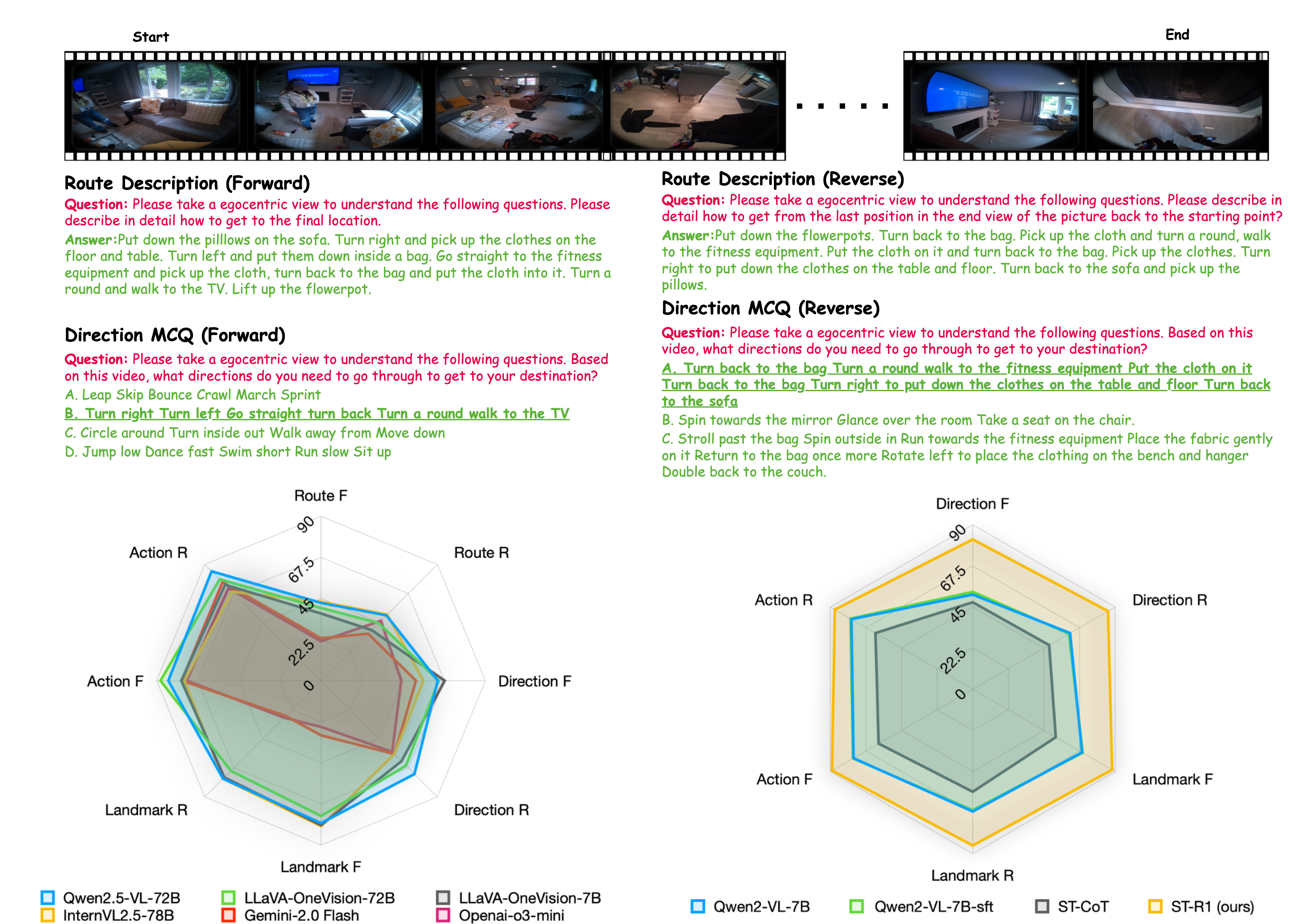}
    \vspace{-0.3em}
    \vspace{-3mm}
        \caption{\textit{Top:}  The Ego-ST bench is characterized by Egocentric \textbf{S}patio-\textbf{T}emporal reasoning questions, and two spatial-temporal reasoning tasks are chosen here as examples: Route Description and Direction change selection.
        \textit{Bottom left:} Shows the performance of each model in a spatial-temporal reasoning task. \textit{Bottom right:} Shows the results of the 4 training methods compared to baseline. (In the radar chart, F stands for forward and R stands for reverse.)}
    \vspace{-1em}
    \label{fig:teaser}
\end{figure*}

\begin{abstract}
Humans excel at spatial-temporal reasoning, effortlessly interpreting dynamic visual events from an egocentric viewpoint. However, whether multimodal large language models (MLLMs) can similarly understand the 4D world remains uncertain. This paper explores multimodal spatial-temporal reasoning from an egocentric perspective, aiming to equip MLLMs with human-like reasoning capabilities. To support this objective, we introduce \textbf{Ego-ST Bench}, a novel benchmark containing over 5,000 question-answer pairs across four categories, systematically evaluating spatial, temporal, and integrated spatial-temporal reasoning. Additionally, we propose \textbf{ST-R1} training paradigm, a video-based reasoning model that incorporates reverse thinking into its reinforcement learning process, significantly enhancing performance. We combine long-chain-of-thought (long-CoT) supervised fine-tuning with Group Relative Policy Optimization (GRPO) reinforcement learning, achieving notable improvements with limited high-quality data. Ego-ST Bench and ST-R1 provide valuable insights and resources for advancing video-based spatial-temporal reasoning research.

\end{abstract}

\section{Introduction}

Multimodal large language models with first-person spatial-temporal reasoning mimic human perception by seamlessly integrating visual, auditory, and textual data. This advanced capability enables a deeper understanding of dynamic environments, thereby supporting highly accurate predictions and context-aware decisions in fields such as autonomous driving, robotics, and augmented reality. Their human-like reasoning fosters more reliable, adaptable, and robust AI systems capable of handling complex tasks. Nevertheless, whether current large language models truly possess such sophisticated spatialtemporal reasoning abilities remains an open question that warrants further investigation.

Recently, this issue has garnered widespread attention. VSI-Bench\cite{yang2024thinking} introduced a benchmark for spatial reasoning and found that spatial reasoning capabilities remain the primary bottleneck for MLLMs. However, it mainly focuses on spatial aspects such as object and room sizes, spatial relationships, and distances, while lacking evaluation data for spatialtemporal reasoning tasks, such as the trajectory of ego motion, environmental description along paths, and action prediction. Moreover, it does not evaluate the network's reverse reasoning abilities. In terms of models, VADAR\cite{marsili2025visual} proposed an agentic approach to decompose spatial reasoning tasks and perform reasoning incrementally, but this approach is limited to static spatial problems and still fails to address spatialtemporal reasoning. Meanwhile, SpacialCoT\cite{liu2025spatialcot} introduced spatial coordinate bi-directional alignment and chain-of-thought spatial grounding to enhance spatial reasoning, yet it still faces challenges in handling spatialtemporal reasoning, such as issues with temporal alignment. Therefore, we require a new benchmark for spatialtemporal reasoning to support research in this area. Moreover, training a model capable of spatialtemporal reasoning remains an important and challenging task.

In this paper, we first propose a novel \textbf{\textit{Ego-ST bench}} designed to evaluate current multimodal large language models and support cutting-edge research on spatialtemporal reasoning from an egocentric perspective. This comprehensive dataset comprises over 5,000 meticulously annotated instances, including detailed question-answer pairs and various multiple-choice questions. Notably, it is the first dataset to annotate both forward and reverse reasoning, thereby facilitating rigorous evaluation and in-depth research on bidirectional reasoning mechanisms. In our Ego-ST bench, our aim is to comprehensively evaluate the advanced perceptual capabilities of MLLMs. Unlike previous benchmarks such as HourVideo~\cite{chandrasegaran2024hourvideo} and VSI-Bench~\cite{yang2024thinking}, our work places a stronger emphasis on accurately describing complete routes, reasoning about directional changes and landmarks, and eliciting more granular and precise model responses. Inspired by Lu's work on LLMs~\cite{lu-etal-2024-rethinking} and the inherent challenge of retracing reasoning paths frequently encountered in everyday human experiences, we introduce the novel concept of \textbf{\textit{Reverse Thinking}} for the first time in the domain of multimodal large-scale model ego-video analysis. This innovative approach encourages MLLMs to adopt a human-centric perspective when interpreting ego-video content, enabling us to rigorously evaluate their capabilities in both spatialtemporal recall and reverse reasoning. Moreover, our benchmark sets a new standard that drives further research and paves the way for significant advancements in this emerging field.

In addition, we propose a new spatial-temporal reasoning training paradigm, \textbf{\textit{ST-R1}}. Building on insights from reverse thinking, the chain-of-thought (CoT) concept~\cite{wei2022chain}, and reinforcement learning approaches exemplified by Deepseek R1~\cite{guo2025deepseek}, we have developed novel strategies for training a robust spatialtemporal reasoning model. Our core idea is to employ reverse thinking as the thought process to enhance spatialtemporal reasoning capabilities. To the best of our knowledge, we are the first to adopt reverse spatialtemporal thinking to boost the spatialtemporal reasoning ability of multimodal models. As shown in Figure~\ref{fig:R1_model}, our ST-R1 Video model restructures the tasks of forward and reverse route descriptions by using the latter as a reasoning process. Specifically, we construct spatialtemporal CoT data that enables the model to learn the reasoning style and logic through Supervised Fine-Tuning (SFT) while retaining key generic capabilities. Subsequently, we employ Group Relative Policy Optimization (GRPO) as a second stage to further enhance the model’s reasoning performance.

On the Ego-ST Bench, we found that existing models still exhibit considerable room for improvement in spatialtemporal reasoning and question answering. Surprisingly, there is no significant performance gap between open-source and closed-source models in this domain; in fact, some open-source models supporting longer context windows even achieve superior performance. This suggests that there is substantial research potential for enhancing the spatialtemporal reasoning capabilities of these networks. By post-training with a small amount of high-quality long CoT data, we developed the ST-R1 model, which significantly outperforms traditional SFT methods. This outcome underscores the effectiveness of our proposed multi-stage post-training strategy.

Our main contributions can be summarized as three-fold:
\begin{itemize}
    \item We introduce a new \textbf{Ego-ST Bench} designed to evaluate multimodal large models from an egocentric perspective. This dataset comprises over 5,000 annotated instances, including both question-answer pairs and multiple-choice questions. Notably, it is the first to annotate both forward and reverse reasoning, thereby supporting comprehensive bidirectional evaluation.
    \item We propose the novel concept of \textbf{Reverse Thinking} in the context of ego-video analysis for multimodal models. This approach mimics human reasoning by retracing the reasoning path, providing a human-centric perspective that enhances spatialtemporal recall and reverse reasoning capabilities—an innovation not previously explored in this domain.
    \item We develop a new spatialtemporal reasoning training paradigm, \textbf{ST-R1}, which leverages insights from reverse thinking, the CoT concept, and reinforcement learning strategies. Our multi-stage training process, starting with SFT using spatialtemporal CoT data and followed by GRPO, significantly improves reasoning performance compared to traditional methods.
\end{itemize}

\section{Related Work}
\textbf{Egocentric Video Understanding.}
Egocentric video understanding is a critical subfield of video analysis that provides a promising avenue for investigating whether multi-modal large language models (MLLMs) can emulate human-like reasoning. Moreover, it offers valuable insights for research in embodied intelligence and robotics. Compared to traditional video understanding tasks—such as those involving MSVD-QA~\cite{10.1145/3123266.3123427}, MSRVTT-QA~\cite{10.1145/3123266.3123427}, ActivityNet-QA~\cite{yu2019activitynet}, and LongVideoBench~\cite{wu2025longvideobench} egocentric video understanding is inherently more complex, demanding greater time and computational resources. In response to these challenges, numerous scholars have introduced several pivotal datasets, including Ego4D~\cite{grauman2022ego4d}, EPIC-KITCHEN~\cite{Damen_2018_ECCV}, HD-EPIC~\cite{perrett2025hd}, and HOI4D~\cite{liu2022hoi4d}. Building upon these resources, researchers have recently begun to explore the capacity of MLLMs in the context of egocentric video, leading to the development of several benchmarks such as HourVideo~\cite{chandrasegaran2025hourvideo}, EgoPlan-Bench2~\cite{qiu2024egoplan}, and X-LeBench~\cite{zhou2025x}. However, none of these benchmarks specifically target the evaluation of higher-level spatial-temporal reasoning capabilities in MLLMs. To address this gap, we propose the Ego-ST bench, which offers a detailed evaluation and analysis framework designed to evaluate these advanced reasoning skills.

\noindent\textbf{Multimodal Spatial Benchmarks.}
In previous studies, numerous scholars have evaluated the spatial capabilities of MLLMs. For instance, SpatialVLM~\cite{chen2024spatialvlm} and SpatialRGPT~\cite{cheng2024spatialrgpt} primarily address 2D image-level spatial understanding. However, spatial-temporal comprehension is critical in the video domain. Although benchmarks such as HourVideo~\cite{chandrasegaran2024hourvideo}, OVO-Bench~\cite{li2025ovo}, and X-LeBench~\cite{zhou2025x} have assessed spatial-temporal understanding, an in-depth evaluation of multimodal spatial-temporal reasoning remains lacking. To address this gap, we propose the Ego-ST bench, which specifically focuses on evaluating higher-level spatial-temporal reasoning abilities in MLLMs.

\noindent\textbf{Mutimodal Reasoning.}
As large language models (LLMs) such as OpenAI’s O1~\cite{o1} and DeepSeek-R1~\cite{guo2025deepseek} continue to gain prominence, a growing body of research has explored their application in multimodal domains. Notable examples include multimodal frameworks like LLava-O1~\cite{xu2024llava} and Open-R1-Multimodal~\cite{open-r1-mutimodal}, which seek to bridge the gap between textual, visual, and other data modalities. However, we contend that the challenges in multimodal domains do not currently require the same chain-of-thought reasoning techniques that have proven effective for mathematics, coding, and logic. Rather, the critical need lies in addressing advanced spatial-temporal reasoning problems that traditional methods struggle to resolve. In response to this gap, we introduce the ST-R1 model, which is specifically designed to excel in complex spatial-temporal reasoning tasks. By leveraging recent advances in multimodal reasoning and focusing on the intricate interplay between spatial and temporal features, the ST-R1 model aims to push the boundaries of multimodal reasoning and establish a new benchmark for tackling these sophisticated challenges.

\begin{figure*}[t!]
    \centering
    \includegraphics[width=\linewidth]{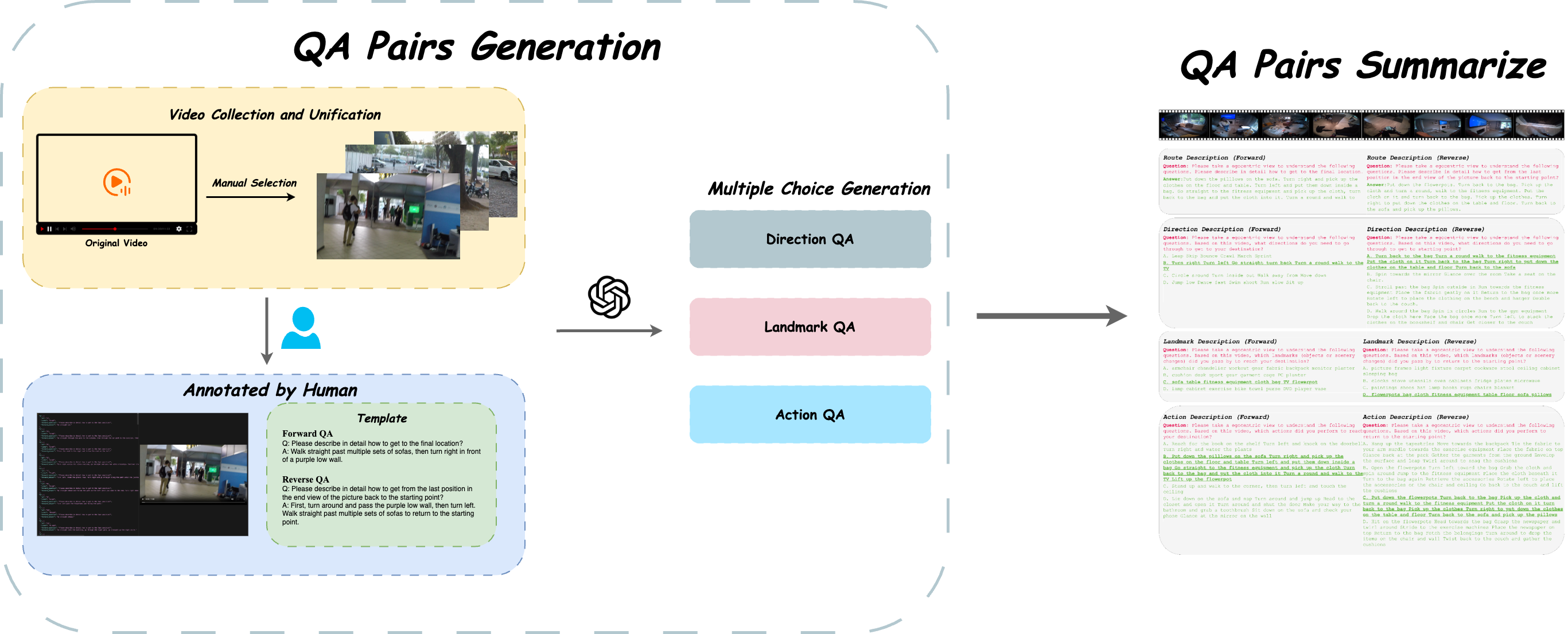}
    \caption{Benchmark pipeline. The pipeline starts by manually filtering segments from different datasets and self-collected video data into a standardised format for consistent processing. Forward and reverse route description QA pairs are then generated through manual annotations and question templates. After obtaining the manually annotated route description QA pairs GPT-4o api is used to generate the corresponding three types of multi-selected QA pairs.}
    \vspace{-0.4cm}
    \label{fig:data_framework}
\end{figure*}

\begin{figure*}[t!]
    \centering
    \includegraphics[width=\linewidth]{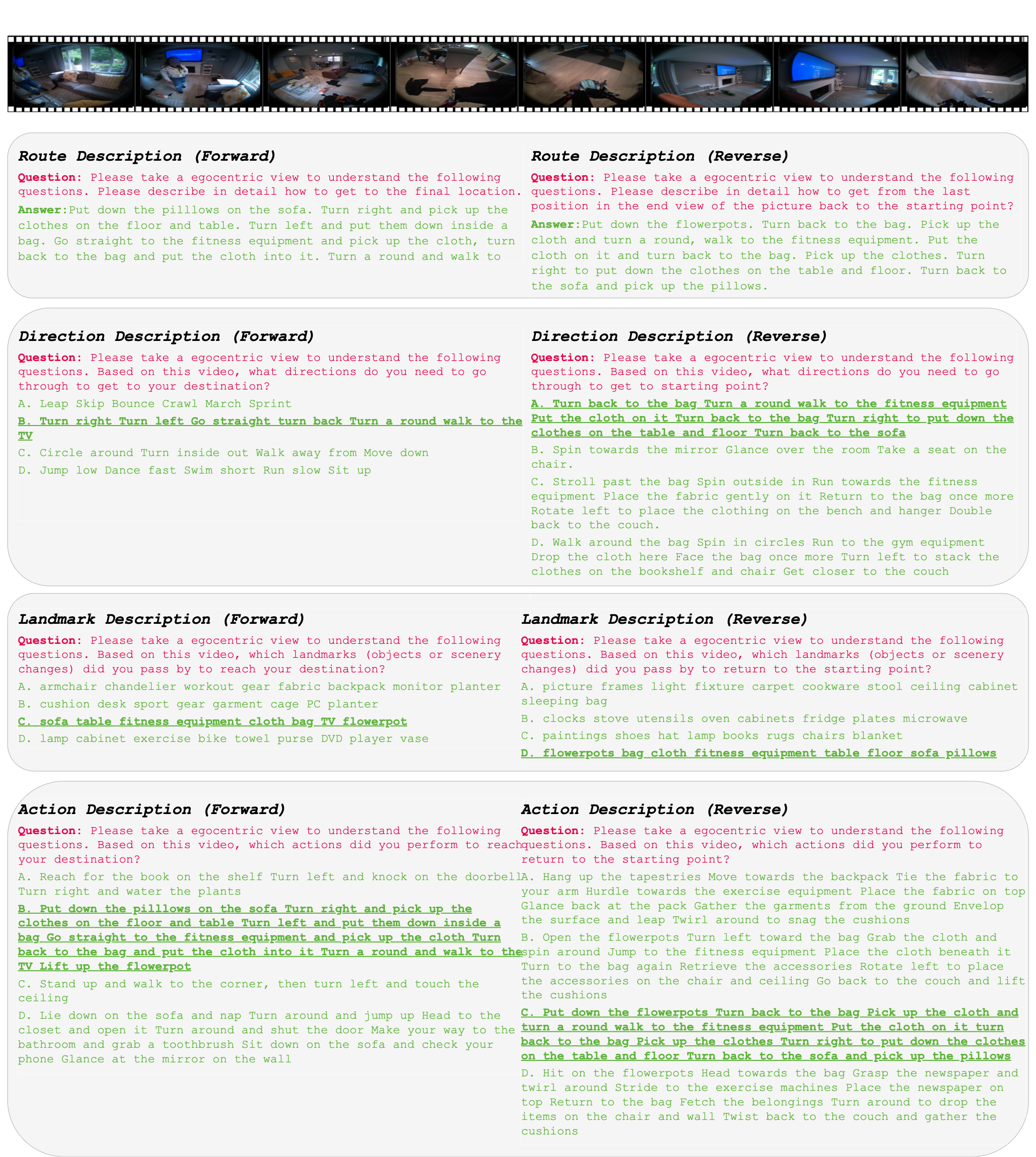}
    \caption{\textbf{Demonstration of tasks for Ego-ST bench.} It includes 8 QAs for 4 types of tasks, which are route description, direction description, landmark description and action description.}
    \vspace{-0.4cm}
    \vspace{-3mm}
    \label{fig:QA_sample}
\end{figure*}

\section{Ego-ST Bench}

\subsection{Overview}
\begin{figure}[t!]
    \centering
    \includegraphics[width=\linewidth]{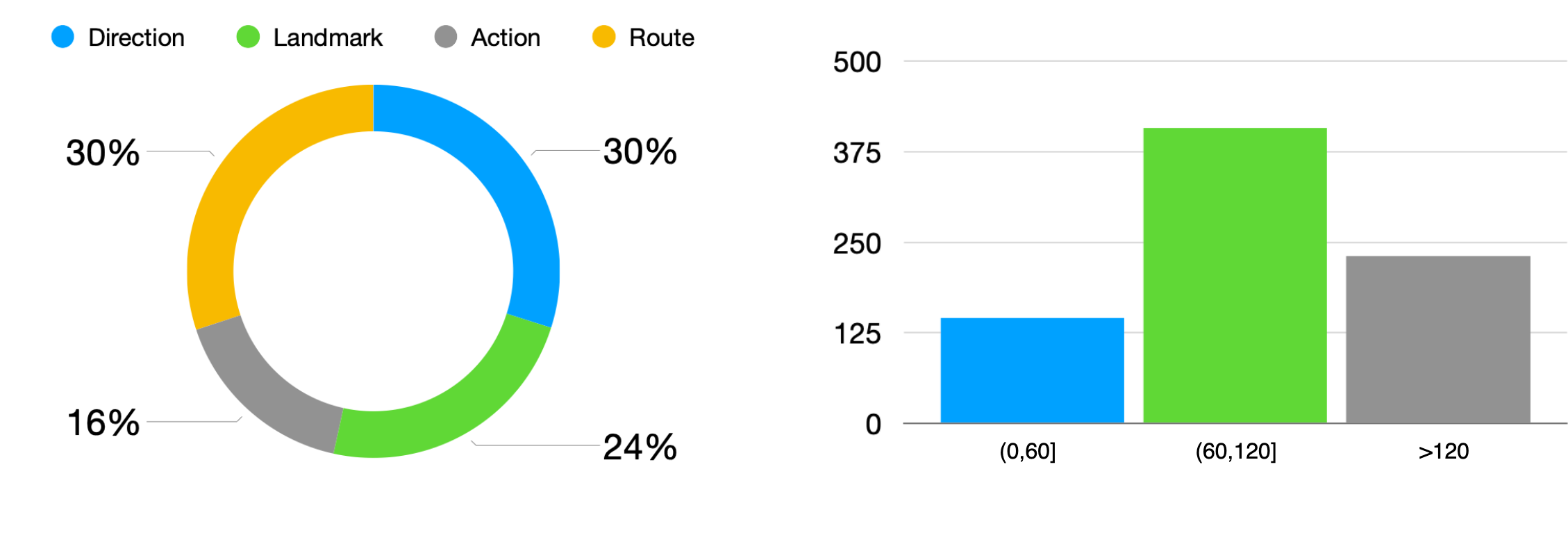}
    \caption{Benchmark Statistics. \textit{Left}: The distribution of tasks across four main categories. \textit{Right}: The video length statistic.}
    \vspace{-0.4cm}
    \label{fig:data_statis}
\end{figure}

We present the Ego-ST bench, which evaluate higher-level spatial-temporal reasoning abilities in egocentric video-based MLLM. The Ego-ST bench consists of over 5000 Question and Answer pairs from 789 real video clips. In total, these videos are divided into two types of scenarios, which include indoor scenarios and ever-changing outdoor complex scenarios. And there are five video sources, including video clips from SUN3D~\cite{xiao2013sun3d}, HUJI~\cite{poleg_wacv16_compactcnn}, DoMSEV~\cite{Silva2018}, Aria Project and self collection video data. The reuse of these existing ego-video datasets provides accurate and difficult annotations. The Ego-ST bench consists of eight tasks of four types: Route Description, Direction Change Description and Action Change Description. Each of these types of tasks contains problem settings for forward and reverse reasoning, fully assessing the capabilities of MLLM in spatial-temporal reasoning. We believe that only a model that truly has the ability to reason inversely about forward video information can be called capable of spatial-temporal reasoning. For an overview of the Ego-ST bench task samples, see Figure~\ref{fig:QA_sample}. For related statistics, please refer to Figure~\ref{fig:data_statis}.

\subsection{Benchmark Construction}

We develop a sophisticated spatial-temporal benchmark construction pipeline to effectively generate high-quality question and answer (QA) pairs at scale, you can see the pipeline framework in Figure~\ref{fig:data_framework}.

\noindent\textbf{Data Collection and Unification.}
We conducted a detailed research and screening of past egocentric video datasets, and finally we selected 6 egocentric video datasets, including: SUN3D~\cite{xiao2013sun3d}, HUJI~\cite{poleg_wacv16_compactcnn} , DoMSEV~\cite{Silva2018}, Aria Everyday Activities~\cite{lv2024aria}, Aria Digital Twin~\cite{Pan_2023_ICCV} and Nymeria~\cite{ma2024nymeria}. In addition to this, we also collect a portion of the video to make up for the lack of richness of the picture in the locations where the original video data was concentrated.
In these data we collated the describable parts by manually filtering and cropping the clips to make sure that the content of each video clip is labellable and has spatial-temporal reasoning implications. In the end, we collated and filtered a total of 789 video clips with videos containing indoor single scenes, multi-scenes and outdoor open complex scenes.

\noindent\textbf{Question-Answer Generation.}
We spent a total of \textbf{\textit{over 150 hours (each person)}} in manual filtering and annotation time, from video data filtering to QA pair creation. We carefully designed and refined the labelling templates for the positive and negative questions for the videos of four different scenarios (2 outdoor scenes and 2 indoor scenes). Among them, as it involves reverse route annotation, it requires the annotator to have a very high concentration and reverse spatial-temporal reasoning ability.

Based on the forward and reverse route annotation information of the video data of the four sections, we elaborated a set of automated multiple-choice question generation and calibration pipeline, as shown in Figure~\ref{fig:data_framework}. Based on this pipeline, we use the gpt-4o to generate three types of multiple-choice questions (direction description, landmark description, and action description) from the annotated data.

\section{Spatial-Temporal R1}
\begin{figure}[t]
    \centering
    \includegraphics[width=\linewidth]{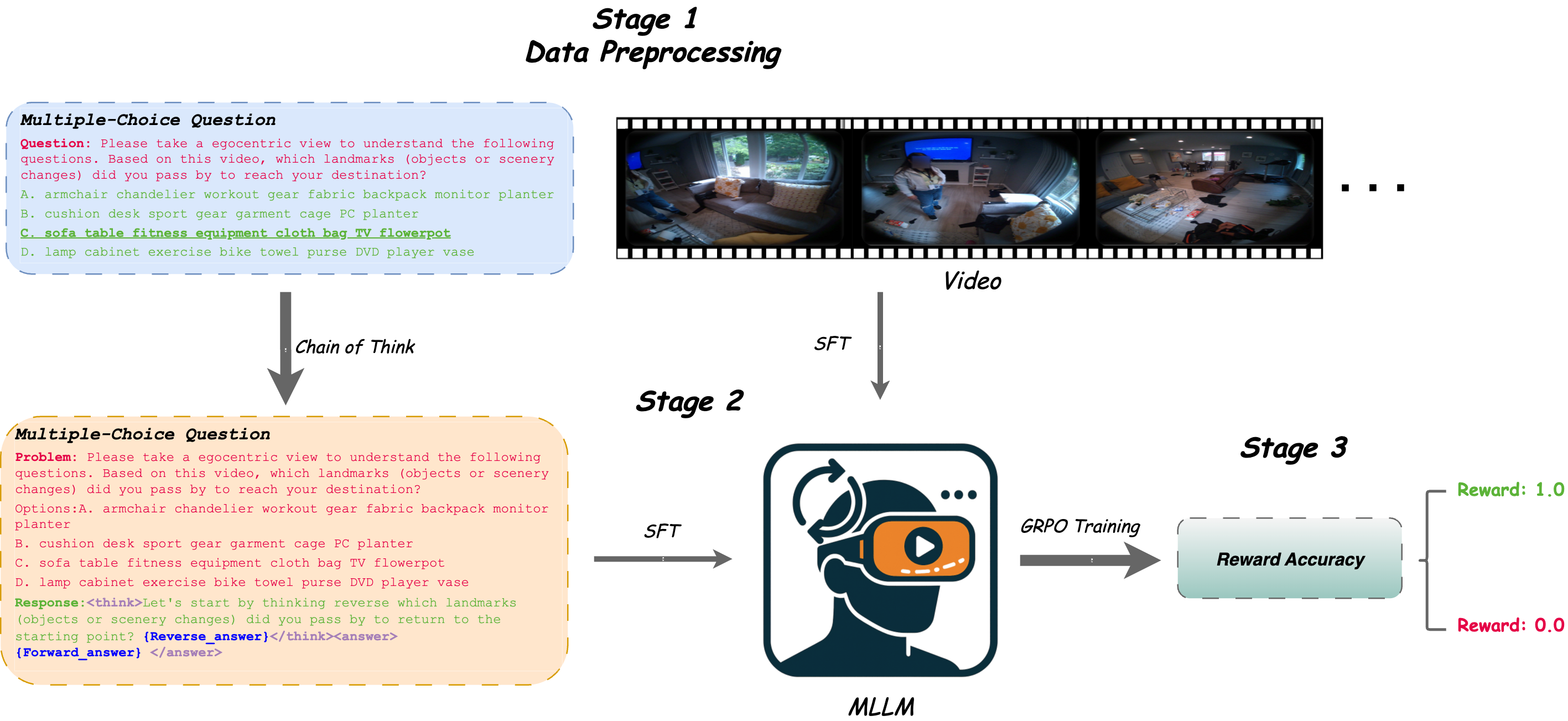}
    \caption{\textbf{Spatial Temporal Reasoning Model.} Our model is trained in two stages: (1) Create Chain of Thought (CoT) data for supervised fine-tuning (SFT). (2) Enhancing the model using the rule-based reinforcement learning GRPO algorithm.}
    \vspace{-0.4cm}
    \label{fig:R1_model}
\end{figure}

Next we will introduce a new paradigm for training in spatial-temporal tasks. The Spatial-Temporal R1 (ST-R1) model is a framework designed to understand and reason about 4D spatial-temporal information in videos. The goal of ST-R1 is to interpret complex events in videos by tracking objects, actions, and their interactions across space and time. By modeling both spatial configurations and temporal dynamics, ST-R1 can comprehend scenarios such as physical events, human activities, and scene changes, enabling it to answer questions or make inferences about video content. Formally, given a video (a sequence of frames) and a query, ST-R1 produces a reasoned answer by analyzing visual cues through time. To achieve this high-level understanding, as shown in Figure~\ref{fig:R1_model} ST-R1 is trained in two stages: first with supervised learning to teach chain-of-thought reasoning and second with reinforcement learning to refine its policy for video question-answering.

\subsection{Chain-of-Thought Supervised Fine-Tuning}
In this stage, the model learns to produce intermediate reasoning steps, not just final answers, when solving video-related questions. 
We prepare a training set $\mathcal{D}_{\text{CoT}} = {(x^{(i)}, c^{(i)}, a^{(i)})}$ where each $x^{(i)}$ is an input (the video content and a question about it), $c^{(i)}$ is a human-crafted chain-of-thought (a step-by-step reasoning process) for that question, and $a^{(i)}$ is the correct answer. The model is fine-tuned to generate the full reasoning trace followed by the answer, effectively learning a reasoning process from the question to the answer. Formally, if we denote the combined reasoning and answer text as $y^{(i)} = (c^{(i)}, a^{(i)})$, the fine-tuning objective maximizes the likelihood of the correct reasoning sequence given the input:

\begin{equation}
\mathcal{L}_{\mathrm{SFT}}(\theta)
= - \mathbb{E}_{(x,c,a)\sim \mathcal{D}_{\mathrm{COT}}}
\bigl[\log P_{\theta}(c,a \mid x)\bigr]
\end{equation}

where $\theta$ are the model parameters. This loss expands over the tokens of the chain-of-thought and answer, ensuring the model learns to think aloud and arrive at the correct answer step by step. We use the forward problem as the thinking process for the reverse problem and the reverse problem as the thinking process for the forward problem.

Integrating forward and reverse reasoning has been shown to improve overall reasoning accuracy\cite{jiang2023forward}, as the model learns not only to derive answers but also to validate them. After this CoT fine-tuning stage, ST-R1 has a strong initial ability to produce reasoned answers for video-based questions, but there is room to optimize its performance and alignment further. This motivates our second training stage using reinforcement learning.

\subsection{Post-training using GRPO}

\textbf{Revisit Group Relative Policy Optimization.} Group Relative Policy Optimization(GRPO) is a recent policy optimization technique tailored for training large models (like language or multimodal models) with complex reasoning tasks. At its core, GRPO is similar in spirit to Proximal Policy Optimization (PPO) but introduces key differences in how the policy is updated and how the reward signal is used. Rather than relying on a learned value function (critic) to estimate expected reward, GRPO entirely foregoes the critic model. Instead, it estimates the baseline or expected reward by sampling a group of outputs from the current policy for each input and using their collective reward statistics. In other words, the model generates multiple candidate answers for a given query or state and uses their relative rewards to decide how to adjust the policy. This group-based advantage estimation removes the need to train a separate value network, simplifying the algorithm and reducing computational overhead.
The development of ST-R1’s training procedure was inspired by the DeepSeek-R1 project, an advanced language reasoning model. DeepSeek-R1\cite{guo2025deepseek} demonstrated that reinforcement learning can be remarkably effective for honing a model’s reasoning abilities, even to the point of learning complex reasoning with minimal supervised data. It highlighted how GRPO can be used effectively in language model training to improve reasoning. The success of DeepSeek-R1 informed our approach for ST-R1: we similarly employ a two-stage training (CoT SFT followed by GRPO-based RL) to develop strong spatial-temporal reasoning capabilities.

\noindent\textbf{GRPO Training for ST-R1.}In the second stage of training, we apply the GRPO algorithm to fine-tune ST-R1 with reinforcement learning. At this point, ST-R1 already has a supervised foundation in reasoning (from the CoT training stage), so the aim is to further improve its accuracy and decision-making on video-based questions by optimizing directly for task-specific rewards. The objective of GRPO training is to adjust the model’s parameters to maximize the expected reward $\mathbb{E}[R(x,y)]$ over the distribution of video questions, using the group-based policy update scheme outlined earlier. ST-R1’s initial policy for this stage $\pi_{\theta_{\text{init}}}$ is set to the CoT fine-tuned model from stage one. We then iteratively improve the policy using GRPO updates: in each iteration, for each training query $x$, multiple answers $y_1,\dots,y_K$ are sampled and scored, and the policy is updated to prefer answers with higher scores, while maintaining closeness to $\pi_{\text{init}}$. 
\begin{equation}
J(\theta) 
= \mathbb{E}_{x}\!\biggl[
  \sum_{k=1}^K w_k \,\log \pi_{\theta}\bigl(y_k \mid x\bigr)
\biggr]
- \beta \, D_{\mathrm{KL}}\!\Bigl(
   \pi_{\theta}(\cdot \mid x)
   \,\Big\|\,
   \pi_{\mathrm{ref}}(\cdot \mid x)
\Bigr)
\end{equation}
Here $w_k$ is the weight assigned to output $y_k$ after reward normalization (for example, $w_k$ could be $\frac{\exp(\tilde{r}k/\tau)}{\sum_j \exp(\tilde{r}j/\tau)}$ for some scaling temperature $\tau$), and the second term is a KL divergence penalty. The KL term, with coefficient $\beta$, measures the divergence between the updated policy $\pi\theta$ and a reference policy $\pi{\text{ref}}$. 

Through these principles, GRPO provides a stable and efficient way to fine-tune ST-R1: it pushes the model toward higher-reward (better reasoning) outputs while maintaining coherence with its initial learned behavior.

\noindent\textbf{Reward Function Design.} A critical aspect in this RL stage is the design of the reward function $R(x,y)$, especially for a complex domain like video understanding. Many of the video understanding tasks are formulated as multiple-choice questions to simplify evaluation. Instead of having the model generate a free-form answer that might be hard to judge automatically, we provide a fixed set of answer options (one correct and the rest incorrect) for a given question. The model’s answer, in this case, is its choice among these options. The reward is then defined in a straightforward way: The model receives a positive reward if it selects the correct option and a zero (or negative) reward if it selects an incorrect option. The multiple-choice formulation thus greatly eases the reward engineering problem for many questions, allowing ST-R1 to learn from explicit right/wrong feedback.

Using the above reward strategies, we train ST-R1 with GRPO by continually generating answers and adjusting the policy. Over many iterations, this process increases the likelihood of the model producing answers that yield high rewards – in other words, answers that are correct and well-reasoned. Due to the multiple-choice reward design, the model receives very clear signals about which answers are right.  The GRPO algorithm’s group-based updates ensure that the model explores different possible answers and learns from comparing them.

\begin{figure*}[t!]
    \captionsetup{type=table}
    \vspace{-0.1cm}
    \centering
    \fontsize{10pt}{10pt}\selectfont
    \setlength\tabcolsep{5pt} 
    \renewcommand{\arraystretch}{1.2} 

\resizebox{\textwidth}{!}{
    {
    \begin{tabular}{r|cc|cccccccc}
        & & & \rotatebox{50}{\(Forward\)} & \rotatebox{50}{\(Reverse\)} & \rotatebox{50}{\(Direct_{\ Forward}\)} & \rotatebox{50}{\(Direct_{\ Reverse}\)} & \rotatebox{50}{\(Landmark_{\ Forward}\)} & \rotatebox{50}{\(Landmark_{\ Reverse}\)} & \rotatebox{50}{\(Action_{\ Forward}\)} & \rotatebox{50}{\(Action_{\ Reverse}\)} \\
        Methods & Avg.(ST) & Avg.(Total) & \multicolumn{2}{c}{\cellcolor{orange!10}Route Description} & \multicolumn{6}{c}{\cellcolor{yellow!10}Multiple-Choice Answer} \\
        \hline
        \rowcolor{navyblue!5}
        \multicolumn{1}{l|}{\textit{Proprietary Models (API)}} &  &  &  &  &  &  &  &  &  &  \\
        Openai-o3-mini & 41.8 & 45.8 & 21.6 & 46.4 & 43.9 & 55.1 & 25.3 & 28.6 & 73.5 & 71.7 \\
        Gemini-2.0 Flash & 42.1 & 47.3 & 23.2 & 36.4 & 52.0  & 56.8  & 29.9 & 27.3 & 76.2 & 76.3 \\
        Gemini-1.5 Pro & 40.4 & 45.2 & 16.0 & 36.2 & 46.7 & 62.8 & 21.5 & 26.3 & 78.1 & 74.0 \\
        \hline
        \rowcolor{navyblue!5}
        \multicolumn{1}{l|}{\textit{Open-source Models (Zero-shot)}} &  &  &  &  &  &  &  &  &  &  \\
        Qwen2-VL-7B & 44.8 & 58.8 & 31.0 & 33.0 & 52.6 & 62.4 & 71.0 & 67.1 & 76.3 & 76.7 \\
        InternVL2.5-8B & 50.6 & 62.2 & 41.8 & 48.0 & 56.7 & 55.7 & 80.9 & 76.1 & 72.5 & 65.8 \\
        Qwen2.5-VL-7B & 49.9 & 62.4 & 37.8 & 43.4 & 54.8 & 63.7 & 75.7 & 70.1 & 78.6 & 74.8 \\
        InternVL2.5-78B & 51.8 & 63.3 & 43.2 & 51.2 & 56.0 & 56.8 & 79.6 & 75.9 & 74.8 & 68.9 \\
        LLaVA-OneVision-7B & 51.5 & 63.9 & 36.8 & 39.2 & 67.7 & 62.3 & 79.1 & 74.9 & 76.5 & 74.3 \\
        LLaVA-OneVision-72B & 53.7 & 65.6 & 39.8 & 44.6 & 64.4 & 65.8 & 74.1 & 69.8 & 87.9 & 78.5 \\
        Qwen2.5-VL-72B & 57.4 & 69.0 & 42.6 & 50.6 & 63.9 & 72.3 & 78.0 & 75.9 & 83.6 & 84.7 \\
    \end{tabular}    }
    }
    \vspace{0.2cm}
    \caption{\textbf{Evaluation on ST-bench.} We designed four categories of questions, each subdivided into forward and reverse reasoning tasks for separate evaluation. The table presents the average scores on the strong temporal reasoning task and the overall average scores for each model.}
    \label{tab:main_table}
    \vspace{-0.4cm}
\end{figure*}

\section{Evaluation on Ego-ST Bench}
\subsection{Evaluation Setup}
\noindent\textbf{Benchmark Models.}
We comprehensively evaluate 10 MLLMs from a diverse model family, including both closed-source and open-source models (the open-source models were tested in large and small-parameter versions, respectively). For proprietary models, we considered OpenAI-o3mini, Google Gemini-2.0, and Gemini-1.5 Pro. For open-source models, we evaluated models from InternVL2.5~\cite{chen2024expanding}, Qwen2-VL~\cite{wang2024qwen2} , Qwen2.5-VL~\cite{bai2025qwen25vltechnicalreport} and LLaVA-OneVision~\cite{li2024llava}. All evaluations were performed in a zero-shot setting and used the default cue for each model. More details of the parameters can be found in the Appendix.

\noindent\textbf{Metric Design.}
Two types of questions are included in our Ego-ST bench, including Open-ended Questions and Multiple-Choice Questions (see Figure~\ref{fig:QA_sample}). Regarding Multiple-Choice Questions (MCQ), we follow standard workflow by using \textit{Accuracy} (\(\mathcal{ACC}\)), based on exact matching, as the primary metric.

\begin{figure}[t!]
    \centering
    \includegraphics[width=\linewidth]{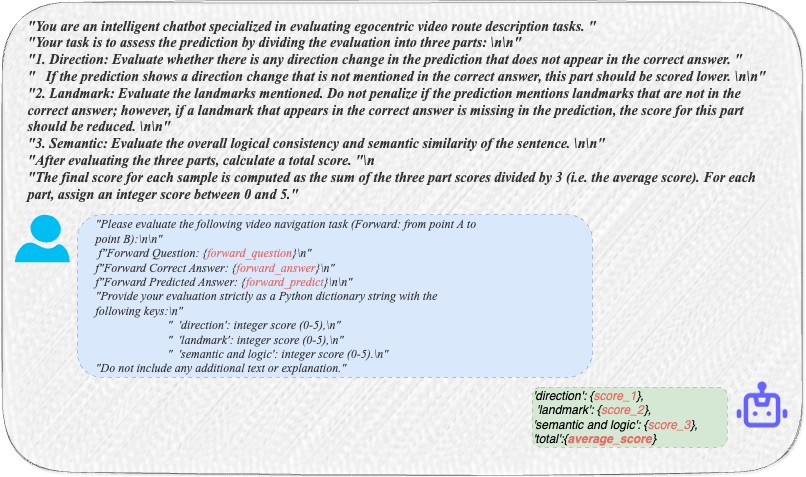}
    \vspace{0.2cm}
    \caption{\textbf{The Prompt for the evaluation.} this prompt is scored in three parts, \textit{Direction:} focuses on MLLM's ability to perceive spatial and temporal changes. \textit{Landmarks:} focuses on MLLM's ability to perceive spatial changes and self-orientation. \textit{Logical Semantics:} focuses on MLLM's ability to organize logical language.}
    \vspace{-0.4cm}
    \label{fig:eval_prompt}
\end{figure}

For the open-ended questions we used standard workflow~\cite{wu2024fmbench,maaz2023video} by using by using LLM judge.We used GPT-4o api and set up a set of spatial-temporal evaluation of route description prompt.The evaluation method we devised for route descriptions is different from the evaluation methods, such as Video-ChatGPT~\cite{maaz2023video}, that use absolute \textit{‘yes’} or \textit{‘no’} and single-score evaluation methods. In total, we consider three aspects of route descriptions, as shown in Figure~\ref{fig:eval_prompt}.
Our approach includes direction changes, landmark changes, and logical semantic consistency. Each section is scored in integer from 0-5, and a final percentage score is calculated for the overall average score.

\begin{table}[t!]
\centering
\renewcommand{\arraystretch}{1.2}

\resizebox{\linewidth}{!}{
\begin{tabular}{cccccc}
\hline
\textbf{Model} & \textbf{Best Part} & \textbf{Avg.} & \textbf{Direction \textit{(D)}} & \textbf{Landmark \textit{(L)}} & \textbf{Semantic \textit{(S)}} \\
\hline
Openai-o3-mini & \textit{S} & 21.6/46.4 & 16.0/46.2 & 24.6/43.1 & 24.0/50.0 \\
Gemini-2.0 Flash & \textit{D} & 23.2/36.4 & 26.6/45.6 & 17.7/23.6 & 25.4/40.0 \\
Gemini-1.5 Pro & \textit{D} & 16.0/36.2 & 18.9/41.4 & 12.0/28.6 & 17.2/38.4 \\
\hline
Qwen2-VL-7B & \textit{L} & 31.0/33.0 & 23.4/27.0 & 41.2/41.2 & 28.0/30.8 \\
Qwen2.5-VL-7B & \textit{L} & 37.8/43.4 & 33.2/40.2 & 46.2/48.2 & 34.2/41.4 \\
Qwen2.5-VL-72B & \textit{L} & 42.6/50.6 & 36.4/49.4 & 52.0/53.4 & 39.0/49.0 \\
InternVL2.5-8B & \textit{L} & 41.8/48.0 & 37.6/44.9 & 48.8/53.2 & 39.0/45.6 \\
InternVL2.5-78B & \textit{L} & 43.2/51.2 & 37.0/49.2 & 51.6/54.0 & 41.0/50.2 \\
LLaVA-OneVision-7B & \textit{L} & 36.8/39.2 & 30.4/34.6 & 46.4/48.0 & 33.4/35.0 \\
LLaVA-OneVision-72B & \textit{L} & 39.8/44.6 & 34.6/43.0 & 48.2/48.8 & 36.4/42.4 \\
\hline
\end{tabular}}
\vspace{0.2cm}
\caption{Detailed scores for route descriptions, which each contain a forward reasoning score and a reverse reasoning score.}
\label{table2}
\vspace{-1cm}
\end{table}

\subsection{Results}

Table~\ref{tab:main_table} shows the overall performance of the model on the Ego-ST bench. Our main observations regarding the spatial-temporal reasoning capabilities of the different models are as follows:

\noindent\textbf{Overall model performance.}
Overall, we have divided into two categories of models, which include Proprietary Models (using commercial api interfaces) and Open Source Models (using local deployments). In addition to evaluating multimodal large language models which support video, we have specifically selected models with strong reasoning capabilities for evaluation (e.g., o3-mini and Gemini-2.0). As shown in Table~\ref{tab:main_table}, we counted the overall scores and the scores of the strong spatial-temporal reasoning task (which includes route descriptions and direction change descriptions) for each model. We can see that under the same test environment and settings, all open source models do not score more than 60 in the strong spatial-temporal reasoning task, but perform very well in both the single spatial comprehension (landmark) and temporal comprehension (action). This reflects that the current multimodal large language modelling frameworks can effectively perceive spatial and temporal single tasks well respectively during training on large amounts of video and image data. From the human point of view, we are always in a 4D environment (3D+time), so the spatial and temporal reasoning and perception ability of the model is something we need to explore and improve in the future.

\noindent\textbf{Mutiple Choice Question.}
We constructed three new types of multiple choice questions based on the open-ended question labels, as shown in Figure~\ref{fig:QA_sample}. The model needs to select the one with the correct chronological order and change among multiple options. These three types of problems cover direction change judgement, landmark scene change judgement and action change judgement. We consider these questions to be spatial-temporal reasoning questions in a wide sense, but each category has a different focus for evaluating the ability of the model. 

\textit{Direction} change is mainly concerned with whether the model has spatial-temporal understanding. \textit{Landmark} changes focus on the ability of the model to recognise spatial semantics. \textit{Action} changes place more emphasis on the temporal detection ability of the model.

As shown in Table~\ref{tab:main_table}, It can be seen that the open-source model perform relatively poorly in direction change, which also indicates that the existing general multimodal large language model lacks of spatial-temporal reasoning ability, but performs relatively well in the lanmark and action, which is also due to the fact that the models nowadays have carried out in-depth research on both spatial and temporal understanding. But when these two task combine together the results are bad.
In addition, based on the overall performance of each model in different tasks, we can see that the increase in the number of model parameters does not lead to a comprehensive improvement in spatial-temporal reasoning ability under the existing multimodal large language model training methods and frameworks. This is the reason why we would like to try to go for the improvement of spatial-temporal reasoning ability of multimodal large language models.

\begin{table}[t!]
\centering
\renewcommand{\arraystretch}{1.2}

\resizebox{\linewidth}{!}{
\begin{tabular}{cccccc}
\hline
\textbf{Model} & \textbf{Training Num.} & \textbf{Avg.} & \textbf{Direction \textit{(D)}} & \textbf{Landmark \textit{(L)}} & \textbf{Action \textit{(A)}} \\
\hline
Qwen2-VL-7B & \textit{0/3582} & 65.4 / 68.4 & 51.8 / 61.5 & 69.1 / 66.9 & 75.4 / 76.7 \\
Qwen2-VL-7B-sft & \textit{630/3582} & 66.2 / 68.1 & 53.3 / 60.8 & 69.5 / 66.2 & 75.7 / 77.3 \\
SFT+GRPO & \textit{630/3582} & 80.3 / 83.1 & 76.4 / 83.7 & 84.0 / 84.4 & 80.4 / 81.1 \\
ST-CoT & \textit{630/3582} & 53.2 / 55.3 & 47.6 / 48.3 & 52.5 / 56.1 & 59.5 / 61.5 \\
ST-R1 & \textit{63/3582} & 64.8 / 63.5 &  54.9 / 53.8  &  70.9 / 71.0 & 68.6 / 65.6 \\
ST-R1 & \textit{126/3582} & 62.4 / 62.1 &  53.0 / 63.5  & 63.4 / 63.0 & 70.7 / 59.8 \\
ST-R1 & \textit{252/3582} & 67.1 / 66.2 & 59.8 / 68.2 & 67.5 / 66.9 & 73.9 / 63.6\\
ST-R1 & \textit{378/3582} & 70.0 / 67.8 & 56.4 / 51.8 & 80.4 / 82.3 & 73.3 / 69.4\\
ST-R1 & \textit{504/3582} & 80.7 / 82.6 & 75.3 / 83.2 & 86.5 / 84.1 & 80.4 / 80.5\\
\textbf{ST-R1} & \textit{630/3582} & \textbf{86.3} / \textbf{86.1} & \textbf{82.0} / \textbf{85.5} & \textbf{88.1} / \textbf{85.5} & \textbf{88.9} / \textbf{87.2 }\\
\hline
\end{tabular}}
\vspace{0.2cm}
  \caption{Training results with small amounts of data under different training paradigms. Each section contains both forward and reverse scores.}
  \label{table3}
\vspace{-1cm}
\end{table}

\noindent\textbf{Open-ended Route Description.}
As illustrated in Table~\ref{table2}, we present detailed scores for each component within the comprehensive route description task. It is evident that the scores for the \textit{direction} and \textit{landmark} categories in the open-ended Route Description task are substantially lower compared to their corresponding scores in the multiple-choice question tasks. By further comparing the differences in performance between the complete route description and three distinct multiple-choice conditions (\textit{direction}, \textit{landmark}, and \textit{action}), we observed that the complete description task poses significantly greater difficulty than the single-dimensional multiple-choice tasks. This disparity highlights that generating complete route descriptions necessitates a more sophisticated integration of overall spatial-temporal information, deeper reasoning regarding temporal causality, and flexible shifts in spatial perspective. Conversely, the multiple-choice tasks predominantly assess recognition and recall of isolated spatial-temporal details. Consequently, the complete route description task, structured as an open-ended question-answering scenario, presents unique challenges that demand higher-order spatial-temporal reasoning capabilities from models.

\noindent Furthermore, model performance in the open-ended description task generally aligns with their performance on the multiple-choice questions. Specifically, across evaluated open-source models, scores in the \textit{Landmark} category under the Spatial Semantic Understanding dimension are consistently the highest. In contrast, performance in the \textit{Direction Changing Environment}, representative of spatial-Temporal Reasoning, remains relatively poor among both open-source and closed-source models. This result aligns with observations from earlier multiple-choice evaluations, underscoring that current multimodal large language model architectures and associated training paradigms exhibit notable shortcomings in effectively handling complex spatial-temporal reasoning tasks.

\begin{figure}[t!]
  \centering
  \begin{minipage}[b]{0.4\textwidth}
    \centering
    \includegraphics[width=\textwidth]{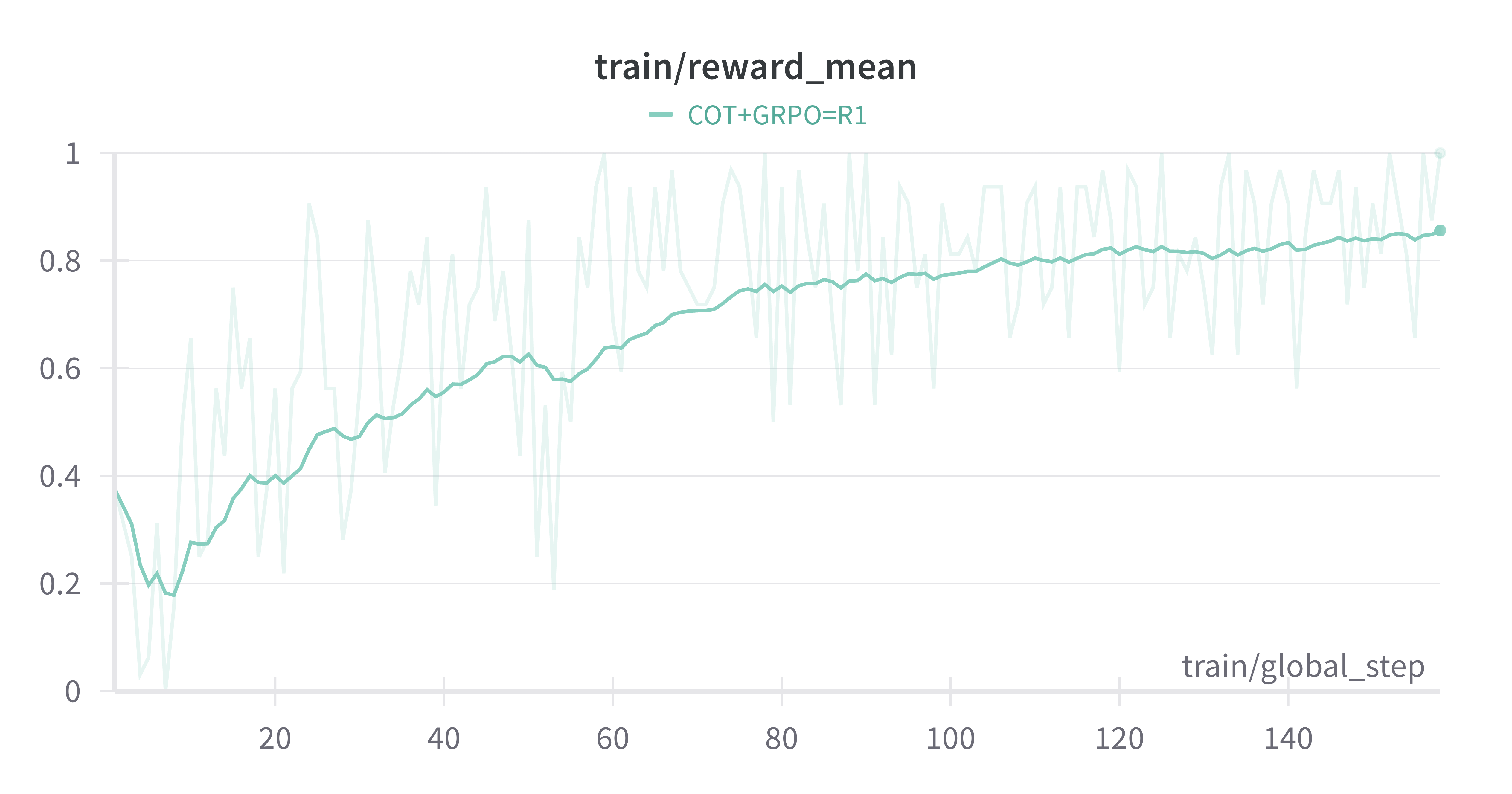}
    \label{fig:image1}
  \end{minipage}
  \hfill
  \begin{minipage}[b]{0.4\textwidth}
    \centering
    \includegraphics[width=\textwidth]{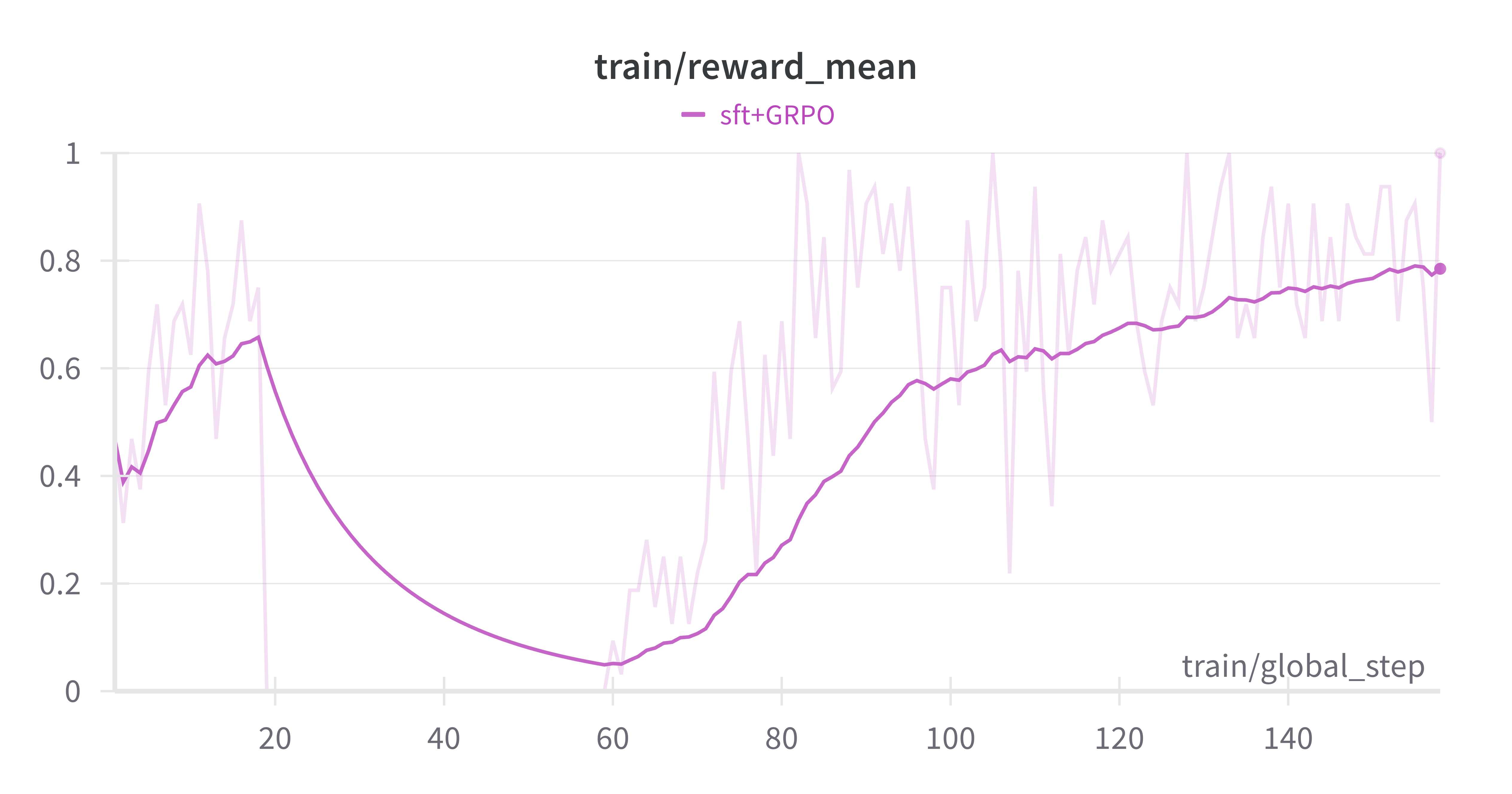}
    \label{fig:image2}
  \end{minipage}
  \caption{\textbf{Left:} Spatial Temporal CoT SFT + GRPO; \textbf{Right:} Regular SFT + GRPO}
  \label{fig:combined}
  \vspace{-0.4cm}
\end{figure}

\subsection{Results of ST-R1 Model}
\textbf{Comparison of different training paradigms.} Based on our thinking about the benchmark evaluation results and Deepseek-R1 training paradigm, we propose a new training paradigm for spatial-temporal reasoning in video. We divide the multiple-choice questions in the overall benchmark into Train and Test in the ratio of 1:6. In addition, we incrementally expanded the training dataset from small to large, aiming to investigate whether our reasoning framework—combining spatial-temporal Chain-of-Thought (CoT) fine-tuning with GRPO reinforcement learning—can achieve superior generalization and spatial-temporal reasoning capabilities, even with limited training data.

As shown in Table~\ref{table3}, the two-stage training approach for the ST-R1 model yields superior generalization performance even with limited training data. Our ST-R1 model, which is post-trained based on Qwen2-VL-7B, achieves significant performance improvements over traditional SFT methods. Initially, the model is trained using spatial-temporal chain-of-thought (CoT) SFT as a cold start, followed by a secondary training phase employing GRPO.  Although direct GRPO tuning improves performance, pre-training with CoT SFT further enhances the results. Experimental results also indicate that when using GRPO for reinforcement alignment, it is crucial to perform prompt-tuning with CoT SFT rather than with conventional SFT. To demonstrate the necessity of spatial-temporal CoT SFT, we also trained a model using regular SFT with GRPO without CoT, as shown in Figure~\ref{fig:combined}. Under the same training conditions, the model trained with regular SFT+GRPO exhibited extremely unstable performance during GRPO training, and its final performance was markedly inferior to that achieved with the CoT SFT+GRPO configuration.

\noindent\textbf{Out of Distribution (OOD) Generalization.}
The experiments in the previous sections showed that a small amount of training data can generalise strongly in same distribution. As shown in Table~\ref{table4}, we continue to experiment with a small amount of training data (630 samples) to test on OOD, and we also validate the performance of \textit{indoor to outdoor} across domains. As we can see from the experimental results, ST-R1 is able to perform well on OOD with a small amount of training data, and ST-R1 achieves a significant improvement of over 32\% in the two types of OOD tasks, respectively. Through the comprehensive analysis of Table~\ref{table3} and Table~\ref{table4}, we can see that ST-R1 is able to show an extremely strong and comprehensive generalisation ability with a small amount of training data.

\begin{table}[t!]
\centering
\renewcommand{\arraystretch}{1.2}

\resizebox{\linewidth}{!}{
\begin{tabular}{cccccc}
\hline
\textbf{Model} & \textbf{Train} & \textbf{Test} & \textbf{Direction \textit{(D)}} & \textbf{Landmark \textit{(L)}} & \textbf{Avg.} \\
\hline
Qwen2-VL-7B & - &  SUN3D  & 45.0 / 49.7  & 73.0 / 73.8 & 59.0 / 61.8\\
ST-R1(indoor-cross) & HUJI+Self+Aria &  SUN3D  &  74.8 / 90.1  & 83.5 / 80.4 & 79.2 / 85.3 \\
Qwen2-VL-7B & - &  HUJI+Self  & 56.6 / 77.0  & 63.9 / 73.9 & 60.3 / 75.5\\
ST-R1(indoor-outdoor) & SUN3D+Aria & HUJI+Self  &  77.6 / 100  & 81.5 / 84.1 & 79.6 / 92.1\\
\hline
\end{tabular}}
\vspace{0.2cm}
  \caption{OOD Testing. We have done detailed tests on indoor scenarios OOD data and \textit{indoor to outdoor} performance migration and generalization.}
  \label{table4}
\vspace{-0.4cm}
\end{table}

\section{Conclusion And Future Work}
In this paper, we explored the spatial-temporal reasoning capabilities of current multimodal large language models, highlighting this as a challenging yet essential research direction. To facilitate comprehensive evaluation and benchmarking, we introduced ST-Bench, a new dataset and benchmark designed specifically to assess existing models’ performance in complex spatial-temporal reasoning tasks. Our results indicate that the performance gap between open-source and closed-source models is minimal. However, we observed a significant performance drop when models are required to organize and generate integrated, open-ended route descriptions, despite their relatively high performance on isolated tasks such as recognizing landmarks or directional changes in multiple-choice settings. This underscores the critical challenge that current multimodal models still lack the ability to effectively integrate complex spatial-temporal information.
Furthermore, we examined optimal post-training strategies for enhancing video-based spatial-temporal reasoning capabilities. Our proposed two-stage post-training approach demonstrated substantial improvements over existing methods, providing valuable insights not only for spatial-temporal reasoning but also for other video reasoning tasks.

\noindent Moving forward, several promising directions merit exploration. These include: (1) enriching multimodal large language models by incorporating additional modalities and diverse data sources to enhance spatial-temporal inference; (2) guiding models toward implicit reconstruction tasks, encouraging a deeper understanding of temporal and spatial dynamics; (3) efficiently developing generalized video reasoning models with robust spatial-temporal reasoning capabilities; and (4) advancing computationally efficient inference methods to support scalable deployment. 

\noindent Addressing these challenges will be critical to advancing the state-of-the-art in multimodal spatial-temporal reasoning.

\section*{Acknowledgements.}
We thank all human annotators and evaluators. This work was mainly supported by OpenInterX Research and the University of Bristol. We would also like to thank the Bristol Digital Futures Institute for providing GPU computing resources.

\medskip
\bibliographystyle{plain}    
\bibliography{main}

\appendix

\end{document}